\documentclass[journal]{IEEEtran}
\IEEEoverridecommandlockouts

\usepackage{tikz}
\usepackage{pdfpages}
\usepackage{cite}
\usepackage{amsmath,amssymb,amsfonts}
\usepackage{algorithmic}
\usepackage{graphicx}
\usepackage{textcomp}
\usepackage{amssymb}
\usepackage{color}
\usepackage{balance}
\usepackage{flushend}
\usepackage{booktabs}
\usepackage[table]{xcolor}
\usepackage[mode=buildnew, subpreambles=true]{standalone}

\usepackage[hyphens]{url}
\usepackage{hyperref}
\hypersetup{colorlinks=false,breaklinks=true} 
\usepackage{cleveref}
\def\BibTeX{{\rm B\kern-.05em{\sc i\kern-.025em b}\kern-.08em
    T\kern-.1667em\lower.7ex\hbox{E}\kern-.125emX}}
    
\begin{document}

% \history{Date of publication xxxx 00, 0000, date of current version xxxx 00, 0000.}
% \doi{10.1109/ACCESS.2020.DOI}

\title{Greener Deep Reinforcement Learning: Analysis of Energy and Carbon Efficiency Across Atari Benchmarks\thanks{This work is supported in part by NSF CPS Grants \#1932300 and \#1931767.}}

%\author{Anonymous Authors}

%%%% commented out for anonymous review %%%%%%%%%
% \thanks{This work is supported in part by NSF CPS Grants \#1932300 and \#1931767.}

\author{\IEEEauthorblockN{Jason Gardner\IEEEauthorrefmark{1}, Ayan Dutta\IEEEauthorrefmark{1}, Swapnoneel Roy\IEEEauthorrefmark{1}, O. Patrick Kreidl\IEEEauthorrefmark{1} and Ladislau Bölöni\IEEEauthorrefmark{2}}\\
\IEEEauthorblockA{\IEEEauthorrefmark{1}University of North Florida, Jacksonville, FL, USA\\
\{n01480000, a.dutta, s.roy, patrick.kreidl\}@unf.edu\\
\IEEEauthorrefmark{2}University of Central Florida, Orlando, FL, USA\\
ladislau.boloni@ucf.edu\\
}
}

\maketitle
\begin{abstract}
The growing computational demands of deep reinforcement learning (DRL) have raised concerns about the environmental and economic costs of training large-scale models. While algorithmic efficiency in terms of learning performance has been extensively studied, the energy requirements, greenhouse gas emissions, and monetary costs of DRL algorithms remain largely unexplored. In this work, we present a systematic benchmarking study of the energy consumption of seven state-of-the-art DRL algorithms, namely DQN, TRPO, A2C, ARS, PPO, RecurrentPPO, and QR-DQN, implemented using Stable Baselines. Each algorithm was trained for one million steps each on ten Atari 2600 games, and power consumption was measured in real-time to estimate total energy usage, CO$_2$-Equivalent emissions, and electricity cost based on the U.S. national average electricity price. Our results reveal substantial variation in energy efficiency and training cost across algorithms, with some achieving comparable performance while consuming up to $24\%$ less energy (ARS vs. DQN), emitting nearly $68\%$ less CO$_2$, and incurring almost $68\%$ lower monetary cost (QR-DQN vs. RecurrentPPO) than less efficient counterparts. We further analyze the trade-offs between learning performance, training time, energy use, and financial cost, highlighting cases where algorithmic choices can mitigate environmental and economic impact without sacrificing learning performance. This study provides actionable insights for developing energy-aware and cost-efficient DRL practices and establishes a foundation for incorporating sustainability considerations into future algorithmic design and evaluation.
\end{abstract}


%\titlepgskip=-15pt
%}

\maketitle
%\IEEEpeerreviewmaketitle

%Our work aims to determine whether there exists a benefit in terms of GHG emissions from using specific algorithms in Reinforcement Learning (RL). Eight RL algorithms were trained to play ten Atari games, and power consumption during training was recorded.

\section{Introduction}
\IEEEPARstart{T}{he} rapid growth of Artificial Intelligence (AI) has led to rising energy demands, contributing significantly to global greenhouse gas (GHG) emissions~\cite{useia}. Information and Communication Technology emissions are estimated to account for $2.1-3.9\%$ of total global emissions~\cite{FREITAG2021100340}, and projections suggest that AI-related energy consumption could more than double by 2030, potentially consuming up to $9\%$ of U.S. electricity generation~\cite{epri}. These trends have spurred growing concern about the environmental and economic sustainability of large-scale AI systems.

Recent examples highlight the enormous energy footprint of training state-of-the-art AI models. Meta’s LLaMA 3.1 auto-regressive language model required an estimated 39.3 million GPU hours, emitting approximately 11,390 tons of CO$_2$-equivalent (tCO$_2$e)~\cite{metallama31card}. Its successor, LLaMA 3.3, added another 2,200 tCO$_2$e over 7 million GPU hours~\cite{metallama33card}. BigScience’s BLOOM-176B large language model (LLM) is estimated to have generated between 24.7 and 50.5 tCO$_2$e during training~\cite{luccioni2022estimatingcarbonfootprintbloom}. Although official GHG emissions for Google’s Gemini and OpenAI’s GPT-4 remain undisclosed, Google reported a 13\% year-over-year increase in corporate GHG emissions in 2023 (14.3 million tCO$_2$e)~\cite{goggle_sustainability}, and Gemini Ultra is estimated to have incurred \$191 million in computation costs~\cite{hai}. OpenAI’s GPT-4 reportedly used around \$78 million in compute~\cite{hai}. Even at inference time, AI systems remain energy-intensive, e.g., a single ChatGPT query requires 2.9 Wh, nearly ten times the energy of a standard Google search (0.3 Wh)~\cite{epri}.

This rising demand is fueled by the growing deployment of energy-hungry hardware. NVIDIA, which holds $80\%$ of the data center GPU market, reported \$60.92 billion in GPU sales in 2024, a $126\%$ increase from 2023. The GPUs sold in 2023 alone are estimated to consume as much electricity as $1.3$ million homes~\cite{tomshardwareSingleModern}. In 2023, $51$ notable AI models were developed in industry, with another $15$ in academia~\cite{hai}, further contributing to AI’s environmental impact.

While much of the attention has focused on large language models (LLMs) and computer vision systems, deep reinforcement learning (DRL) represents another class of AI algorithms with high computational and energy requirements. DRL models often require millions of environment interactions and prolonged training on GPUs, which can result in substantial power consumption and associated CO$_2$ emissions. However, unlike LLMs, the energy footprint of DRL algorithms has received relatively little attention. More importantly, there has been no systematic effort to compare different DRL algorithms in terms of energy consumption, greenhouse gas emissions, or monetary cost.

In this work, we present the first comprehensive benchmarking study of the energy requirements of eight state-of-the-art DRL algorithms, DQN, TRPO, A2C, ARS, (Recurrent)PPO, and QR-DQN, implemented using \textit{Stable Baselines}\footnote{https://stable-baselines3.readthedocs.io/en/master/guide/algos.html}. These algorithms were trained for one million steps on ten Atari 2600 games while recording real-time power consumption. From these measurements, we estimate total energy usage, CO$_2$-equivalent emissions, and the electricity cost based on the U.S. national average price. Our results reveal substantial variability in energy efficiency and training cost across algorithms, providing actionable insights for developing energy-aware and cost-efficient DRL practices. This study lays the groundwork for incorporating sustainability considerations into the design and evaluation of future DRL algorithms.

%Our paper is structured as follows.

\section{Related Work}

\subsection{Environmental Impact of Machine Learning}

The rapid growth of machine learning models has raised concerns regarding their environmental and monetary costs. Early work by \cite{strubell2020energy} and \cite{patterson2022carbon} quantified the substantial carbon emissions of training large neural models, sparking the ``Green AI'' initiative proposed by \cite{schwartz2020green}, which advocated reporting energy consumption as a standard evaluation metric. Other studies, such as \cite{wu2022sustainable}, extended this perspective by suggesting standardized protocols for energy and hardware efficiency reporting. Despite these efforts, most RL research still focuses predominantly on performance improvements, neglecting computational sustainability.

\subsection{Reinforcement Learning Efficiency and Reproducibility}

In the context of DRL, reproducibility and computational efficiency were first systematically examined by \cite{henderson2018deep}, who identified high variance and hyperparameter-sensitive behavior in DRL training. The need for standardized evaluation has since been emphasized in works such as \cite{engstrom2020implementation} and \cite{islam2017reproducibility}. More recently, \cite{henderson2020towards} highlighted the necessity of energy and carbon reporting for RL, but comprehensive empirical studies remain rare.

\subsection{Algorithmic Advances in Deep Reinforcement Learning}

Most DRL algorithmic advances have prioritized performance and sample efficiency. The introduction of DQN by \cite{mnih2015human} and its distributional variants, such as QR-DQN \cite{dabney2018distributional}, significantly improved value estimation in Atari benchmarks. Actor-critic methods, including A2C and A3C \cite{mnih2016asynchronous}, reduced variance and training time by leveraging parallelism, while policy-gradient improvements such as PPO \cite{DBLP:journals/corr/SchulmanWDRK17}, TRPO \cite{schulman2015trust}, and RecurrentPPO \cite{DBLP:journals/corr/SchulmanWDRK17} offered better stability. ARS \cite{NEURIPS2018_7634EA65}, although originally proposed for continuous control, gained attention as a computationally inexpensive baseline due to its simple linear policy updates. Distributed frameworks, such as IMPALA \cite{espeholt2018impala} and scalable PPO \cite{cobbe2021phasic}, further optimized throughput, but energy cost reporting was largely omitted.

\subsection{Positioning of Our Work}

To the best of our knowledge, no prior study has provided a systematic, large-scale comparison of DRL algorithms in terms of energy efficiency, carbon emissions, and monetary cost across multiple games. While \cite{schwartz2020green} and \cite{henderson2020towards} outlined the importance of such measurements, empirical investigations remain limited. Our work bridges this gap by evaluating eight DRL algorithms spanning value-based, actor-critic, and evolutionary approaches under both sparse and dense reward settings. We present aggregated statistics, cost-efficiency trade-offs, and outlier analyses to provide actionable insights for environmentally sustainable DRL research.

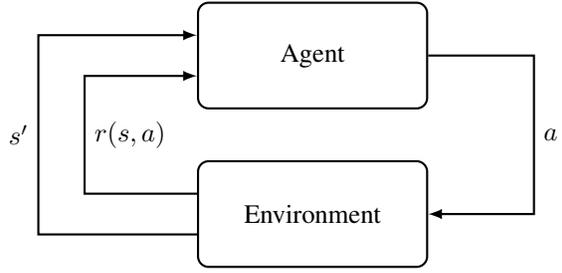
\begin{figure}
\tikzstyle{block} = [rectangle, draw, 
    text width=8em, text centered, rounded corners, minimum height=4em]
    
\tikzstyle{line} = [draw, -latex]
\begin{tikzpicture}[node distance = 6em, auto, thick]
    \node [block] (Agent) {Agent};
    \node [block, below of=Agent] (Environment) {Environment};
    
     \path [line] (Agent.0) --++ (4em,0em) |- node [near start]{$a$} (Environment.0);
     \path [line] (Environment.190) --++ (-6em,0em) |- node [near start] {$s'$} (Agent.170);
     \path [line] (Environment.170) --++ (-4.25em,0em) |- node [near start, right] {$r(s,a)$} (Agent.190);
\end{tikzpicture}
  \caption{Illustration of Reinforcement~Learning. Variables $s$ and $a$ denote the current state of the world and the action taken by the agent in that state, respectively, while $s'$ is the state to which the agent transitions as a result of that action, and $r(s,a)$ is the associated reward.}
  \label{fig:RL_illust}
\end{figure}

\section{Background}
Classical Reinforcement Learning (RL) focuses on learning optimal policies through trial-and-error interactions between an agent and its environment (see Fig. \ref{fig:RL_illust}), typically using tabular methods such as Q-learning or policy iteration. However, these methods struggle with high-dimensional state and action spaces, which has motivated the shift toward Deep Reinforcement Learning (DRL)~\cite{mnih2015human}, where deep neural networks enable scalable function approximation and allow RL to tackle complex tasks such as Atari games~\cite{van2016deep}, robotics~\cite{kulbaka2024gdm}, and autonomous driving~\cite{kiran2021deep}. Deep reinforcement learning leverages deep neural networks to approximate policies or value functions, enabling agents to act optimally in high-dimensional environments. The algorithms benchmarked in this work -- DQN, TRPO, A2C, ARS, (Recurrent)PPO, and QR-DQN -- represent different methodological families:  
(1) \textit{Value-based methods} learn an action-value function \(Q(s,a)\) or its distribution, selecting actions by maximizing estimated returns.  
(2) \textit{Policy gradient methods} directly optimize a parameterized policy \(\pi_\theta(a|s)\) via gradient ascent on expected rewards.  
(3) \textit{Actor-Critic methods} combine both approaches by using a policy (actor) guided by a learned value function (critic).  
(4) \textit{Derivative-free methods}, such as ARS, use random search in the policy parameter space rather than gradient-based updates.  

Policies can be either \textit{stochastic}, where \(\pi(a|s)\) defines a probability distribution over actions, or \textit{deterministic}, where a single action \(a = \mu(s)\) is selected for each state $s$. Algorithms also differ in their data usage:  
\textit{On-policy methods} learn only from data collected using the current policy, ensuring stable updates but requiring frequent sampling, whereas \textit{off-policy methods} can reuse past experience from replay buffers or other policies, improving sample efficiency at the cost of potential instability.

The following subsections summarize the working principles of each algorithm along with key mathematical formulations.

\subsection{Deep Q-Network (DQN)}
The Deep Q-Network (DQN)~\cite{mnih2015human} approximates the optimal action-value function $Q^*(s,a)$ using a deep neural network parameterized by $\theta$:
\begin{equation*}
Q_\theta(s,a) \approx Q^*(s,a) = \mathbb{E}[r + \gamma \max_{a'} Q^*(s',a') | s_t=s, a_t=a].
\end{equation*}
Training minimizes the temporal-difference (TD) loss:
\[
\mathcal{L}(\theta) = \mathbb{E}_{(s,a,r,s') \sim \mathcal{D}} \Big[ \big( y^{\text{DQN}} - Q_\theta(s,a) \big)^2 \Big],
\]
where
\[
y^{\text{DQN}} = r + \gamma \max_{a'} Q_{\theta^-}(s',a')
\]
and $\theta^-$ are target network parameters updated periodically. DQN uses experience replay from buffer $\mathcal{D}$ to decorrelate training samples and stabilize learning.

% \subsection{Double Deep Q-Network (DDQN)}
% DQN tends to overestimate action values due to the \(\max\) operator. Double Deep Q-Network (DDQN)~\cite{van2016deep} mitigates this by decoupling action selection and evaluation:
% \[
% y^{\text{DDQN}} = r + \gamma Q_{\theta^-} \left( s', \arg\max_{a'} Q_{\theta}(s',a') \right).
% \]
% This results in more stable learning and improved policy performance. DDQN was selected to examine whether such algorithmic refinements translate to improved energy efficiency.

% % \subsection{Deep Deterministic Policy Gradient (DDPG)}
% % Deep Deterministic Policy Gradient (DDPG)~\cite{lillicrap2015continuous} is an off-policy actor-critic method designed for continuous control. The actor network \(\mu_\phi(s)\) outputs deterministic actions, while the critic estimates \(Q_\theta(s,a)\). The critic minimizes the Bellman loss:
% % \[
% % \mathcal{L}(\theta) = \mathbb{E}\Big[ \big( r + \gamma Q_{\theta^-}(s', \mu_{\phi^-}(s')) - Q_\theta(s,a) \big)^2 \Big],
% % \]
% % and the actor is updated via the deterministic policy gradient:
% % \[
% % \nabla_\phi J(\phi) = \mathbb{E}_{s \sim \mathcal{D}} \Big[ \nabla_a Q_\theta(s,a)\big|_{a=\mu_\phi(s)} \nabla_\phi \mu_\phi(s) \Big].
% % \]
% % Although designed for continuous actions, DDPG is included to represent the family of off-policy actor-critic methods.

\subsection{Trust Region Policy Optimization (TRPO)}
Trust Region Policy Optimization (TRPO)~\cite{schulman2015trust} is an on-policy policy-gradient method that constrains policy updates to ensure monotonic policy improvement. The optimization problem is:
\[
\max_\theta \, \mathbb{E}_{s,a \sim \pi_{\theta_{\text{old}}}} \left[ \frac{\pi_\theta(a|s)}{\pi_{\theta_{\text{old}}}(a|s)} \, \hat{A}(s,a) \right]
\]
subject to:
\[
\mathbb{E}_s \left[ D_{\mathrm{KL}} \big( \pi_{\theta_{\text{old}}}(\cdot|s) \,||\, \pi_\theta(\cdot|s) \big) \right] \leq \delta.
\]
Here, \(\hat{A}(s,a)\) is the advantage estimate, and \(\delta\) controls the size of the trust region. TRPO uses conjugate gradient and line search to solve this constrained optimization efficiently. It serves as a canonical baseline for stable on-policy policy optimization.

% \subsection{Asynchronous Advantage Actor-Critic (A3C)}
% Asynchronous Advantage Actor-Critic (A3C)~\cite{mnih2016asynchronous} improves training stability by asynchronously running multiple actor-learners in parallel. Each worker updates a shared global network using:
% \[
% \nabla_\theta J(\theta) = \mathbb{E}\Big[ \nabla_\theta \log \pi_\theta(a|s) \hat{A}(s,a) + \beta \nabla_\theta H(\pi_\theta(\cdot|s)) \Big],
% \]
% where \(H(\pi)\) is an entropy term encouraging exploration. The advantage is computed as:
% \[
% \hat{A}(s_t,a_t) = \sum_{i=0}^{k-1} \gamma^i r_{t+i} + \gamma^k V_\theta(s_{t+k}) - V_\theta(s_t).
% \]
% A3C was chosen for its historically significant scalability and reduced wall-clock training time.

\subsection{Augmented Random Search (ARS)}
Augmented Random Search (ARS)~\cite{NEURIPS2018_7634EA65} is a derivative-free optimization method. Given a linear policy \(\pi_\theta(s)\), ARS perturbs the policy parameters with random noise vectors \(\delta_i\) and updates as:
\[
\theta_{t+1} = \theta_t + \frac{\alpha}{b\sigma} \sum_{i=1}^b \left[ R(\theta_t + \nu \delta_i) - R(\theta_t - \nu \delta_i) \right] \delta_i,
\]
where \(R(\cdot)\) is the episodic  return, \(\nu\) the noise scale, $\alpha$ is the learning rate, and $\sigma$ the standard deviation of the $b$ rollout returns. ARS is computationally lightweight, making it an important baseline for evaluating whether simple methods yield energy savings.

% \subsection{Proximal Policy Optimization (PPO)}
% Proximal Policy Optimization (PPO)~\cite{DBLP:journals/corr/SchulmanWDRK17} simplifies TRPO by using a clipped surrogate objective:
% \[
% \mathcal{L}^{\text{CLIP}}(\theta) = \mathbb{E}\left[ \min \left( r_t(\theta)\hat{A}_t,\; \text{clip}\big(r_t(\theta), 1-\epsilon,1+\epsilon\big)\hat{A}_t \right) \right],
% \]
% where \(r_t(\theta) = \frac{\pi_\theta(a_t|s_t)}{\pi_{\theta_{\text{old}}}(a_t|s_t)}\). PPO is widely adopted for its stability and ease of implementation, making it a key benchmark for energy-performance trade-offs.

\subsection{Proximal Policy Optimization (PPO)}

Proximal Policy Optimization (PPO)~\cite{DBLP:journals/corr/SchulmanWDRK17} is a widely adopted on-policy actor-critic algorithm that improves policy stability by constraining updates within a trust region using a clipped surrogate objective. The optimization objective is expressed as
\begin{equation}
L^{\text{CLIP}}(\theta) = \mathbb{E}_t \left[ \min \left( r_t(\theta) \hat{A}_t, \; \text{clip}(r_t(\theta), 1-\epsilon, 1+\epsilon) \hat{A}_t \right) \right],
\end{equation}
where \(r_t(\theta) = \frac{\pi_\theta(a_t|s_t)}{\pi_{\theta_{\text{old}}}(a_t|s_t)}\) is the probability ratio between the new and old policies, \(\hat{A}_t\) is the advantage estimate, and \(\epsilon\) is a hyperparameter controlling the trust region. This formulation prevents excessively large policy updates, improving learning stability compared to vanilla policy gradient methods.

In this work, we specifically utilize the standard and the Recurrent PPO (RecurrentPPO) baselines from Stable Baselines 3 (SB3) Contrib library, which extend the standard PPO implementation by adding support for recurrent policies. A Long Short-Term Memory (LSTM) network is integrated into the policy architecture ({\em MlpLstmPolicy}) to capture temporal dependencies in partially observable environments. Other than this recurrent extension, the algorithm’s behavior, including optimization procedure, clipping mechanism, and advantage estimation, remains identical to SB3’s core PPO implementation. PPO was chosen due to its proven sample efficiency, robustness, and status as a standard baseline in DRL research.

\subsection{Quantile Regression DQN (QR-DQN)}
QR-DQN~\cite{dabney2018distributional} extends DQN by modeling the distribution of returns. QR-DQN approximates the Q-function distribution using N quantiles ${Z_i(s,a)}_{i=1}^N$ at fixed quantile levels $\tau_i = (i-0.5)/N$. The quantile regression loss is:
\[
\mathcal{L}(\theta) = \frac{1}{N} \sum_{i=1}^N \sum_{j=1}^N \rho_{\tau_i}^\kappa \left( y_j - Z_i(s,a) \right),
\]
where $y_j = r + \gamma Z_j(s', \arg\max_{a'} \bar{Q}(s',a')), \bar{Q}(s,a) = \frac{1}{N}\sum_{i=1}^N Z_i(s,a)$, and $\rho_\tau^\kappa(u) = |\tau - \mathbf{1}_{u<0}| \cdot L_\kappa(u)$ is the quantile Huber function with 
\[L_\kappa(u) = \begin{cases} \frac{1}{2}u^2 & \text{if } |u| \leq \kappa \\ \kappa(|u| - \frac{1}{2}\kappa) & \text{otherwise.} \end{cases}\]
QR-DQN was included to evaluate whether the added complexity of distributional RL significantly impacts energy consumption.

%%%%%%%%%%%% A2C %%%%%%%%%%%%%%%%%%%%%%%%
\subsection{Advantage Actor-Critic (A2C)}
Advantage Actor-Critic (A2C) is a synchronous variant of the A3C algorithm~\cite{mnih2016asynchronous} that eliminates the non-determinism of asynchronous updates using batched, synchronous gradient computation across multiple parallel environments. The policy (actor) parameters $\theta_{\pi}$ is updated by maximizing the expected advantage-weighted log-likelihood:
\[
\nabla_{\theta_\pi} J(\theta_\pi) = 
\mathbb{E}\left[ \nabla_{\theta_\pi} \log \pi_{\theta_\pi}(a|s)\, \hat{A}(s,a) + \beta \nabla_{\theta_\pi} H(\pi_{\theta_\pi}(\cdot|s)) \right],
\]
where \(H(\pi)\) is an entropy term (weighted by \(\beta\)) to encourage exploration.  
The advantage function is estimated using the temporal difference error as:
\[
\hat{A}(s_t, a_t) = r_t + \gamma V_{\theta_v}(s_{t+1}) - V_{\theta_v}(s_t),
\]
and the critic with parameters $\theta_v$ is updated by minimizing the squared value loss:
\[
\mathcal{L}_V(\theta_v) = \left( r_t + \gamma V_{\theta_v}(s_{t+1}) - V_{\theta_v}(s_t) \right)^2.
\]
A2C is computationally more efficient than A3C while retaining similar performance~\footnote{https://openai.com/index/openai-baselines-acktr-a2c/}, making it a practical baseline for energy consumption analysis.

% \begin{table}[h]
% \centering
% \caption{Comparison of DRL algorithms benchmarked in this study.}
% \begin{tabular}{|lcccc|}
% \hline
% \textbf{Algorithm} & \textbf{Type} & \textbf{Policy} & \textbf{On/Off-Policy} & \textbf{Complexity} \\
% \hline
% DQN & Value-based & Implicit (via Q-values) & Off-policy & Moderate \\
% %DDQN & Value-based & Implicit (via Q-values) & Off-policy & Moderate \\
% %DDPG & Actor-Critic & Deterministic policy & Off-policy & High \\
% TRPO & Policy gradient & Stochastic policy & On-policy & High \\
% %A3C & Actor-Critic & Stochastic policy & On-policy & High (parallelized) \\
% A2C & Advantage Actor-Critic & Policy-gradient & On-policy %using synchronous advantage estimation to reduce variance and improve stability compared to vanilla policy gradient methods. 
% & High \\%Moderate sample efficiency; higher computational cost \\%due to parallel environments. \\
% ARS & Derivative-free & Linear policy & On-policy & Low \\
% PPO & Actor-Critic & Stochastic policy & On-policy & Moderate \\
% QR-DQN & Distributional value-based & Implicit (quantiles) & Off-policy & High \\
% \hline
% \end{tabular}
% \label{tab:algo_comparison}
% \end{table}

\section{Methodology}

All experiments were conducted on a workstation equipped with an Intel Xeon W-2245 CPU, 128\,GB of system memory, a 512\,GB NVMe system disk, and an NVIDIA RTX A5000 GPU with 24\,GB of GDDR6 memory. No other training load was simultaneously run on the machine. This setup allows us to accurately and consistently compare the performance of the algorithms. %This hardware configuration was selected to provide consistent performance across all experiments and to reflect a typical high-end research laboratory setup for deep reinforcement learning training.

The experiments utilized the Stable Baselines 3 (SB3)~\cite{stable-baselines3} and SB3 Contrib libraries to ensure standardized and reproducible implementations of state-of-the-art RL algorithms. Additional software components included Gymnasium 0.29.1~\cite{kwiatkowski2024gymnasiumstandardinterfacereinforcement} as the simulation environment, Tensorflow 2.16.1~\cite{tensorflow2015-whitepaper}, and NVIDIA CUDA 12.1.105~\cite{10.1145/1401132.1401152} for GPU acceleration. CodeCarbon 2.4.2~\cite{codecarbon} was employed to measure power consumption through its Intel Running Average Power Limit (RAPL) integration and derivative pyRAPL implementation. GPU power consumption was monitored using NVIDIA’s \texttt{nvidia-smi}. Carbon intensity data for electricity consumption were obtained from the Electricity Maps API (v3)~\cite{electricitymaps}. 

Eight reinforcement learning algorithms were benchmarked: DQN, TRPO, A2C, ARS, PPO, RecurrentPPO, and QR-DQN. These algorithms were selected because they represent diverse methodological families, including value-based, policy-gradient, actor-critic, and distributional approaches, and they are commonly used as benchmarks in DRL research. The default SB3 policy implementations were used for all algorithms, with {\em MlpPolicy} (a multilayer perceptron) for most models and {\em MlpLstmPolicy} (LSTM-based) for RecurrentPPO, to ensure consistency with standard experimental practices.

Atari 2600 games were chosen as the experimental environments due to their established role as reproducible benchmarks for DRL~\cite{bellemare13arcade,reproduce}. The selected games, i.e., Asteroids, Beam Rider, Boxing, Breakout, Centipede, Chopper Command, Ms. Pac-Man, Pong, Space Invaders, and Video Pinball, cover a range of game dynamics and difficulty levels while maintaining discrete action spaces and low input dimensionality, making them computationally efficient. Each environment used the \textit{NoFrameSkip-v4} variant, and input frames were preprocessed by converting to grayscale and resizing to 84$\times$84 pixels using the WarpFrame wrapper from SB3, reducing computational load while preserving relevant spatial information.

Each algorithm was trained for one million steps per game. Model checkpoints and energy measurements were recorded every 10,000 steps to align with SB3's logging and checkpointing frequency. Power consumption was measured separately for the CPU, RAM, and GPU, and then aggregated to compute the total energy consumption. Carbon emissions were estimated using the real-time carbon intensity from the Electricity Maps API\footnote{https://www.electricitymaps.com/}. Training costs were calculated using both the U.S. national average electricity rate (\$0.1401/kWh) from the Bureau of Labor Statistics~\cite{bls} and local utility rates from the [name hidden] Energy Authority (\$0.11006/kWh)~\cite{jea_namehidden}, providing a range of realistic energy cost scenarios.

To enable a unified comparison of algorithm efficiency, we normalized performance and dividing the average episodic reward achieved by each model by its total energy consumption. This yielded a single metric, \textit{Average Normalized Performance per Kilowatt Hour} (NPpkWh), allowing direct comparison of energy-performance trade-offs among algorithms. The detailed per-game results are presented in Tables~\crefrange{table:asteroids}{table:videopinball}, and a summary comparison across all games is provided in Table~\ref{table:summary}.

\section{Results and Discussion}

Table~\ref{table:summary} presents the Average Normalized Performance per Kilowatt Hour (NPpkWh) for all seven deep reinforcement learning algorithms benchmarked across ten Atari environments. ARS achieved the highest NPpkWh ($0.08142$), nearly $4.50$x more efficient than the least efficient algorithm, QR-DQN ($0.01889$). TRPO and PPO ranked below ARS, with NPpkWh scores of $0.04838$ and $0.03977$, respectively, outperforming DQN, Recurrent PPO, and QR-DQN by approximately $1.3$-$2.5$x.

\begin{table}[!ht]
\centering
\includegraphics[width=0.9\linewidth]{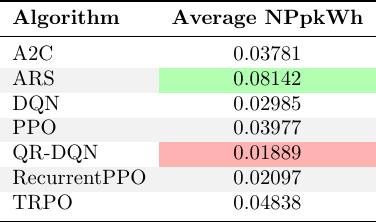}
\caption{Normalized Performance per Kilowatt Hour (higher is better). The highest and lowest numbers are highlighted in green and red, respectively.}
\label{table:summary}
\end{table}

%%%%%%%%%%%%%%% Convergence table -- might need to change later %%%%%%%%%%
% \begin{table}[!ht]
% \centering
% \includegraphics[width=\linewidth]{plots/convergence_table.pdf}
% \caption{Convergence (Number of Steps)}
% \label{table:convergence}
% \end{table}

\subsection*{Energy Efficiency and Algorithmic Trade-offs}
The substantial energy efficiency of ARS stems from its derivative-free optimization strategy, which avoids computationally expensive forward and backward passes through deep networks. By directly perturbing policy parameters, ARS minimizes GPU utilization, memory access, and replay buffer operations, which significantly lowers energy usage. Despite originating in continuous-control tasks, ARS adapted well to the structured observations and dense reward signals of Atari games.

However, ARS's performance may not generalize to domains characterized by sparse rewards, high-dimensional sensory input, or partial observability, where gradient-based methods better exploit small reward signals through backpropagation. Thus, while ARS represents an energy-efficient solution in well-structured domains, its applicability to real-world, high-complexity tasks remains limited.

Among the gradient-based methods, TRPO and PPO achieved strong NPpkWh values due to improved sample efficiency and training stability. TRPO's trust-region policy updates constrain parameter changes, preventing destabilizing updates and reducing wasted computation. Interestingly, PPO outperformed RecurrentPPO in this metric. This suggests that in Atari, where stacked frames already provide sufficient state information, the added recurrence introduces computational overhead and potential instability without offering proportional performance gains.  
%RecurrentPPO, which differs from SB3’s standard PPO only in adding LSTM-based recurrent policies, achieved convergence by effectively modeling temporal dependencies and managing partial observability, thereby reducing redundant training iterations. 
While both algorithms incur higher per-update computational costs, their superior sample efficiency led to fewer total updates and thus lower overall power consumption.

Conversely, DQN and QR-DQN suffered from high replay buffer overheads, and A2C’s synchronous update mechanism required more frequent gradient evaluations, leading to poor energy efficiency ($0.02985, 0.01889, 0.03781$ NPpkWh, respectively). These findings highlight that sample efficiency and stable learning dynamics, rather than per-step computational simplicity, are the primary drivers of energy efficiency. While NPpkWh provides an aggregated measure of energy efficiency, it does not reveal how quickly different algorithms achieve high performance. Since training duration directly impacts both total energy consumption and electricity costs, we further examine the relationship between training time and mean episodic reward to contextualize the observed efficiency differences better.

\subsection*{Training Time vs. Mean Reward}
\begin{figure}[!ht]
\centering
\includegraphics[width=\linewidth]{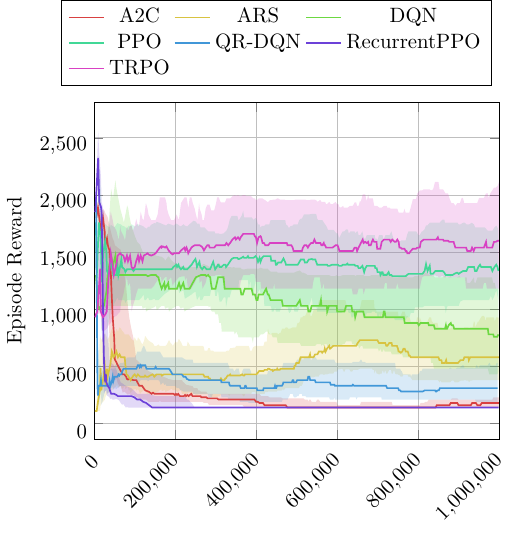}
\caption{Environment: Asteroids\\Rolling Median (window=25\%) + IQR (25-75\%)}
\label{figure:asteroidsiqr}
\end{figure}

\begin{figure}[!ht]
\centering
\includegraphics[width=\linewidth]{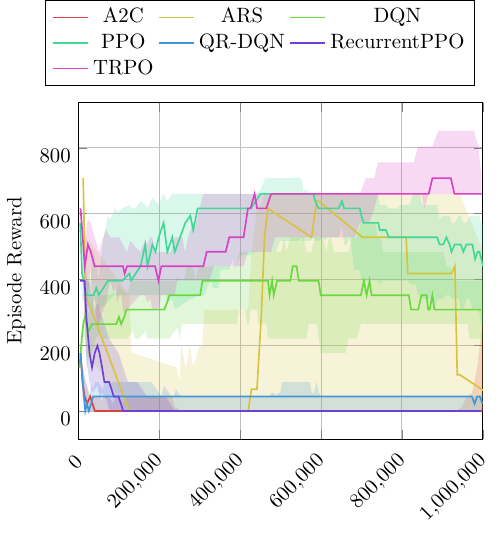}
\caption{Environment: Beam Rider\\Rolling Median (window=25\%) + IQR (25-75\%)}
\label{figure:beamrideriqr}
\end{figure}

\begin{figure}[!ht]
\centering
\includegraphics[width=\linewidth]{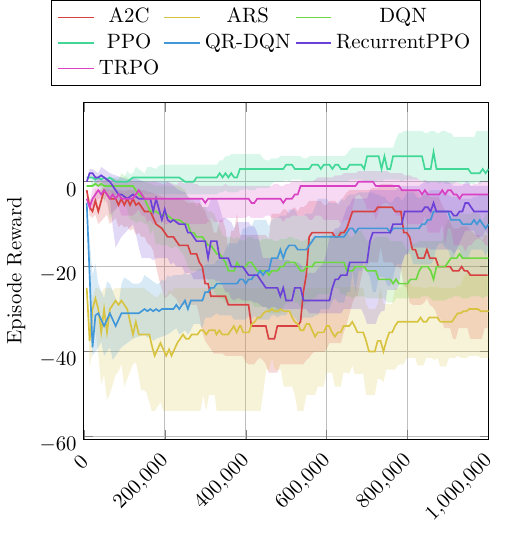}
\caption{Environment: Boxing\\Rolling Median (window=25\%) + IQR (25-75\%)}
\label{figure:boxingiqr}
\end{figure}

\begin{figure}[!ht]
\centering
\includegraphics[width=\linewidth]{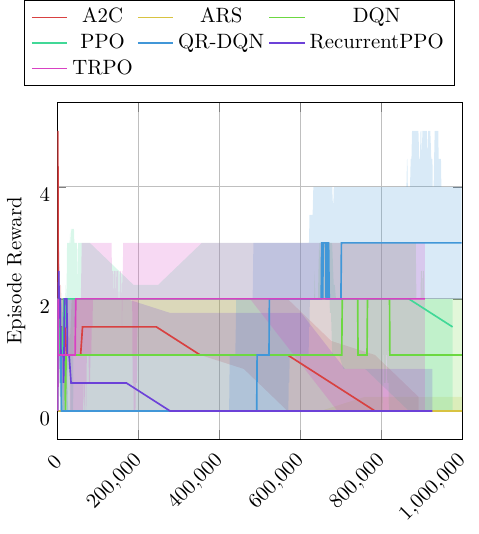}
\caption{Environment: Breakout\\Rolling Median (window=25\%) + IQR (25-75\%)}
\label{figure:breakoutiqr}
\end{figure}

\begin{figure}[!ht]
\centering
\includegraphics[width=\linewidth]{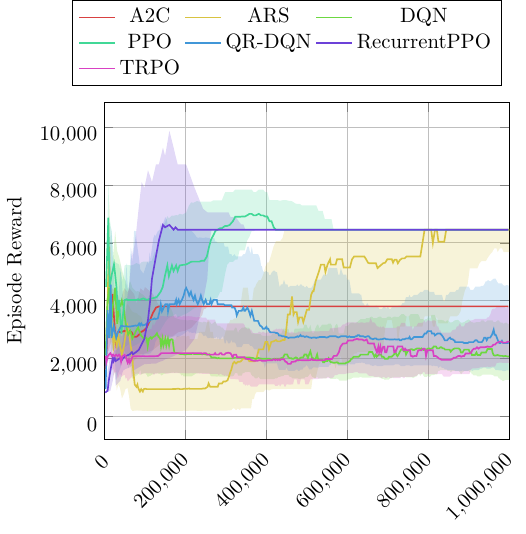}
\caption{Environment: Centipede\\Rolling Median (window=25\%) + IQR (25-75\%)}
\label{figure:centipedeiqr}
\end{figure}

\begin{figure}[!ht]
\centering
\includegraphics[width=\linewidth]{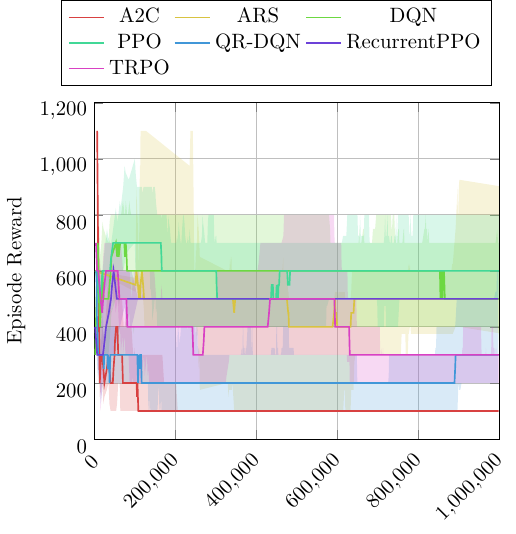}
\caption{Environment: Chopper Command\\Rolling Median (window=25\%) + IQR (25-75\%)}
\label{figure:choppercommandiqr}
\end{figure}

\begin{figure}[!ht]
\centering
\includegraphics[width=\linewidth]{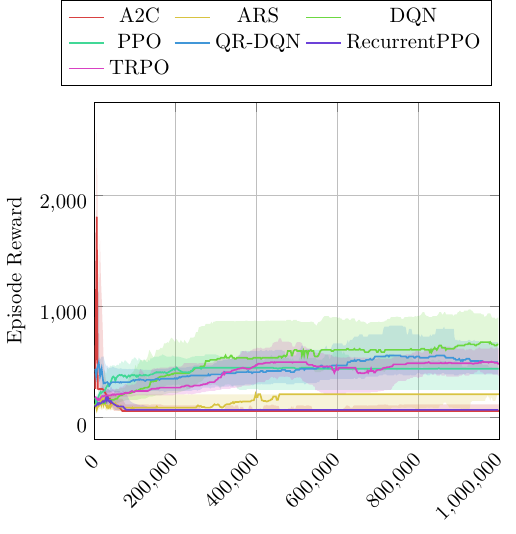}
\caption{Environment: Ms. Pacman\\Rolling Median (window=25\%) + IQR (25-75\%)}
\label{figure:mspacmaniqr}
\end{figure}

\begin{figure}[!ht]
\centering
\includegraphics[width=\linewidth]{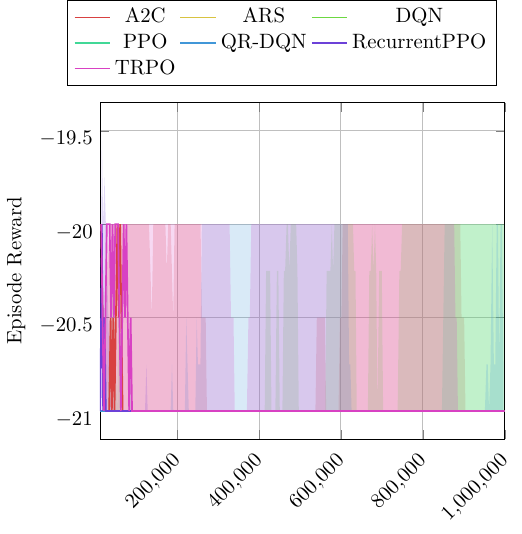}
\caption{Environment: Pong.\\
  Rolling Median (window=25\%) + IQR (25-75\%)}
\label{figure:pongiqr}
\end{figure}

\begin{figure}[!ht]
\centering
\includegraphics[width=\linewidth]{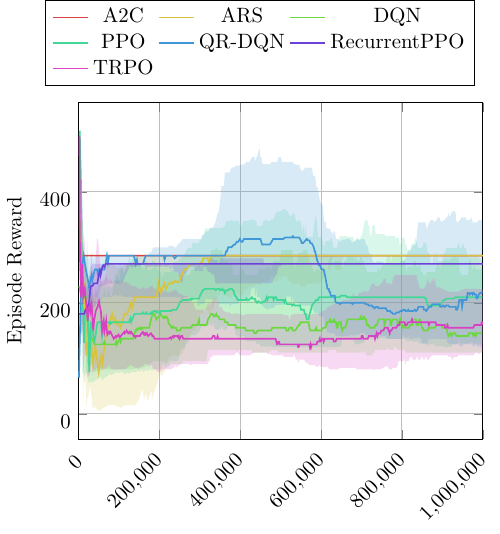}
\caption{Environment: Space Invaders\\Rolling Median (window=25\%) + IQR (25-75\%)}
\label{figure:spaceinvadersiqr}
\end{figure}

\begin{figure}[!ht]
\centering
\includegraphics[width=\linewidth]{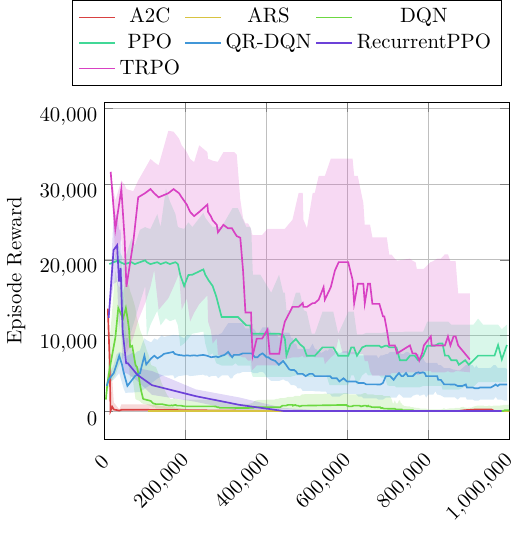}
\caption{Environment: Video Pinball\\Rolling Median (window=25\%) + IQR (25-75\%)}
\label{figure:videopinballiqr}
\end{figure}

\begin{figure}[!ht]
\centering
\includegraphics[width=\linewidth]{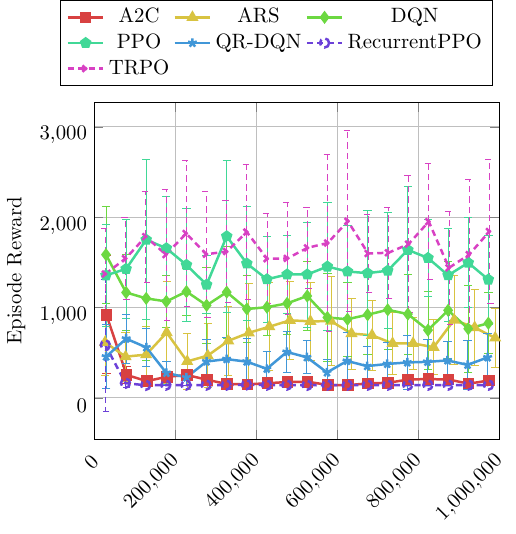}
\caption{Environment: Asteroids\\Binned (20) mean ± std}
\label{figure:asteroidsbinned}
\end{figure}

\begin{figure}[!ht]
\centering
\includegraphics[width=\linewidth]{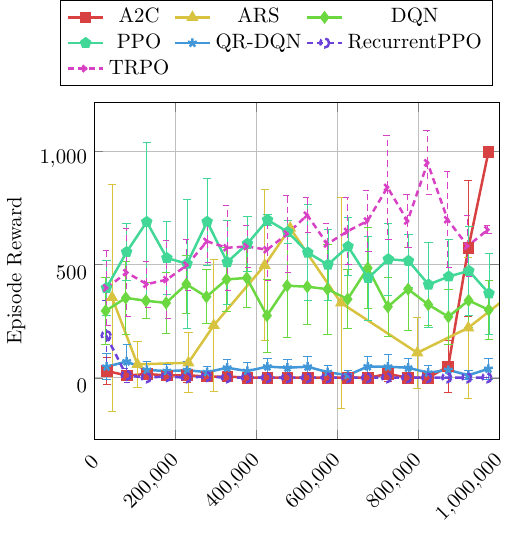}
\caption{Environment: Beam Rider\\Binned (20) mean ± std}
\label{figure:beamriderbinned}
\end{figure}

\begin{figure}[!ht]
\centering
\includegraphics[width=\linewidth]{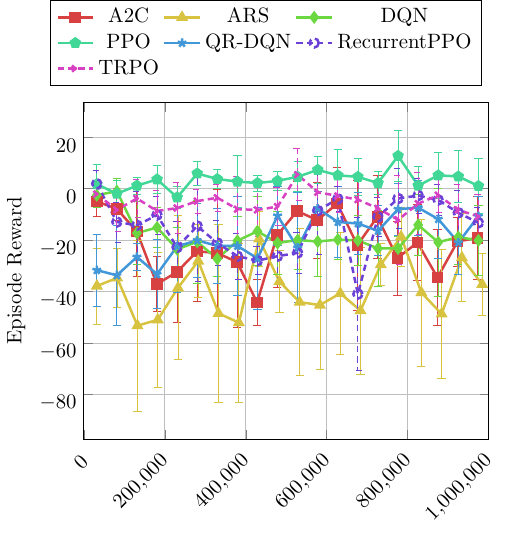}
\caption{Environment: Boxing\\Binned (20) mean ± std}
\label{figure:boxingbinned}
\end{figure}

\begin{figure}[!ht]
\centering
\includegraphics[width=\linewidth]{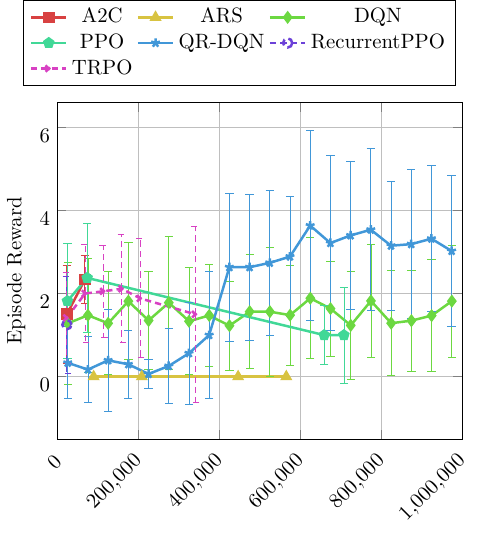}
\caption{Environment: Breakout\\Binned (20) mean ± std}
\label{figure:breakoutbinned}
\end{figure}

\begin{figure}[!ht]
\centering
\includegraphics[width=\linewidth]{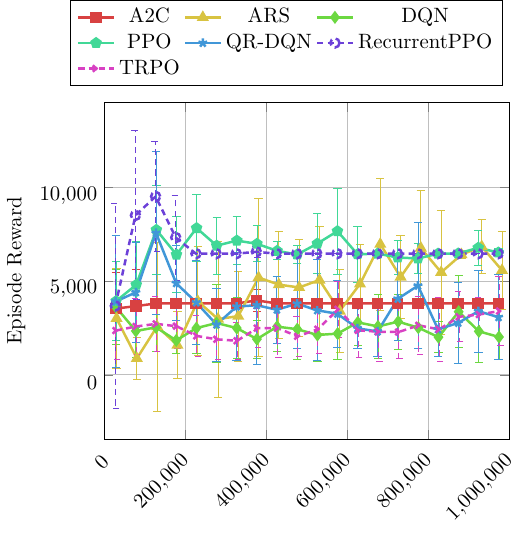}
\caption{Environment: Centipede\\Binned (20) mean ± std}
\label{figure:centipedebinned}
\end{figure}

\begin{figure}[!ht]
\centering
\includegraphics[width=\linewidth]{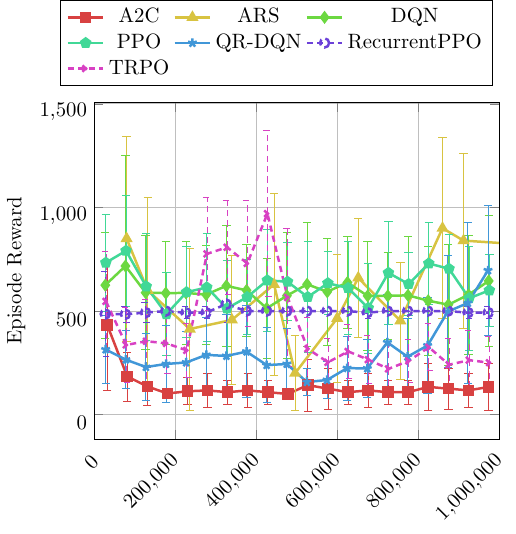}
\caption{Environment: Chopper Command\\Binned (20) mean ± std}
\label{figure:choppercommandbinned}
\end{figure}

\begin{figure}[!ht]
\centering
\includegraphics[width=\linewidth]{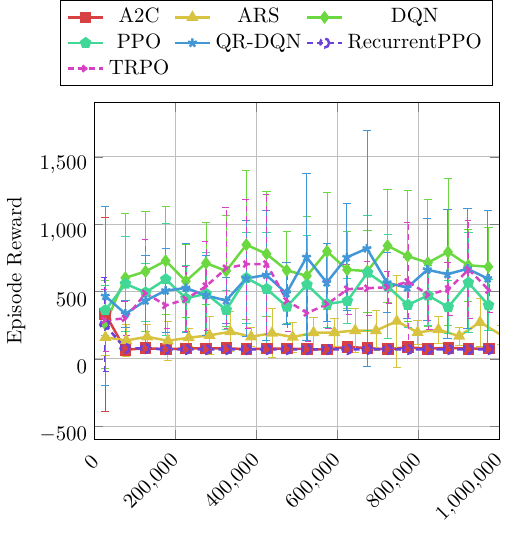}
\caption{Environment: Ms. Pacman\\Binned (20) mean ± std}
\label{figure:mspacmanbinned}
\end{figure}

\begin{figure}[!ht]
\centering
\includegraphics[width=\linewidth]{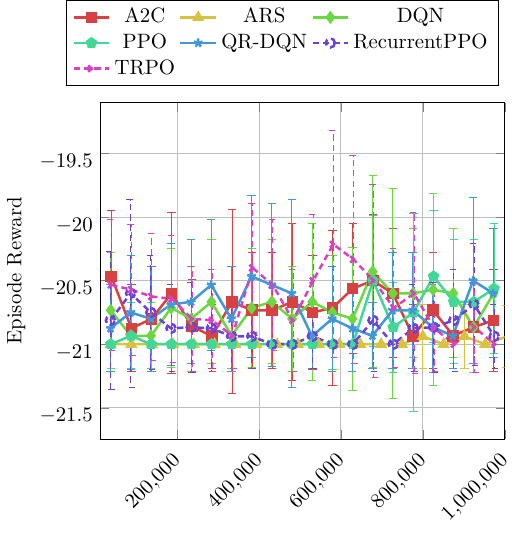}
\caption{Environment: Pong\\Binned (20) mean ± std}
\label{figure:pongbinned}
\end{figure}

\begin{figure}[!ht]
\centering
\includegraphics[width=\linewidth]{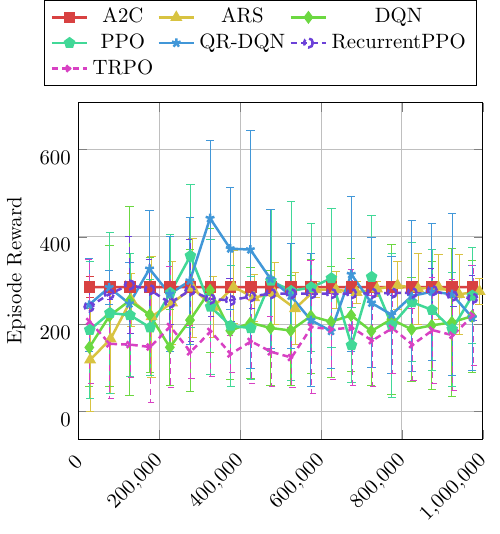}
\caption{Environment: Space Invaders\\Binned (20) mean ± std}
\label{figure:spaceinvadersbinned}
\end{figure}

\begin{figure}[!ht]
\centering
\includegraphics[width=\linewidth]{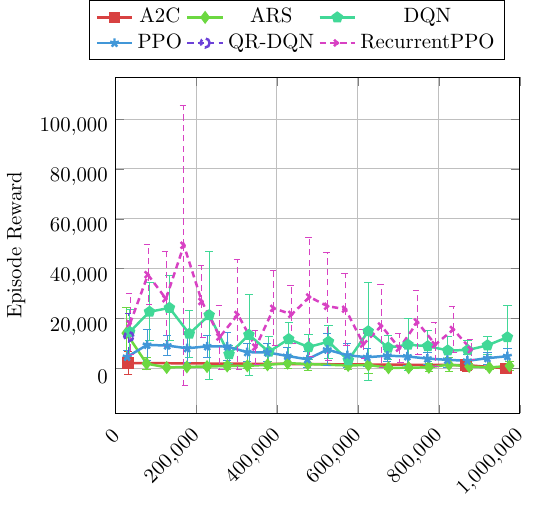}
\caption{Environment: Video Pinball\\Binned (20) mean ± std}
\label{figure:videopinballbinned}
\end{figure}

\begin{figure}[!ht]
\centering
\includegraphics[width=\linewidth]{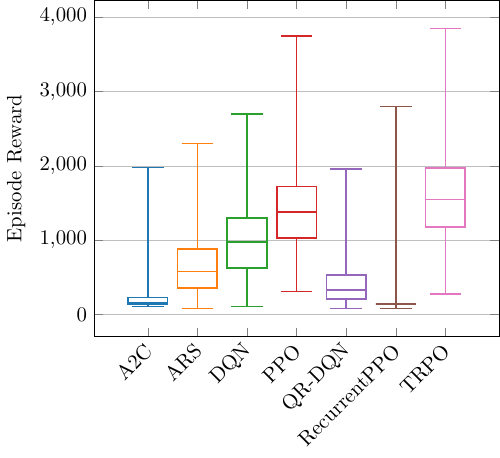}
\caption{Environment: Asteroids\\Reward Distribution}
\label{figure:asteroidsdist}
\end{figure}

\begin{figure}[!ht]
\centering
\includegraphics[width=\linewidth]{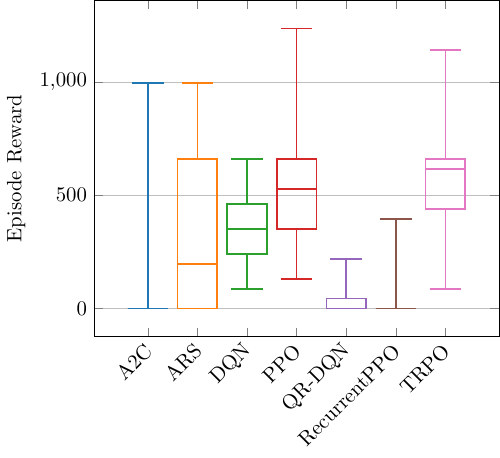}
\caption{Environment: Beam Rider\\Reward Distribution}
\label{figure:beamriderdist}
\end{figure}

\begin{figure}[!ht]
\centering
\includegraphics[width=\linewidth]{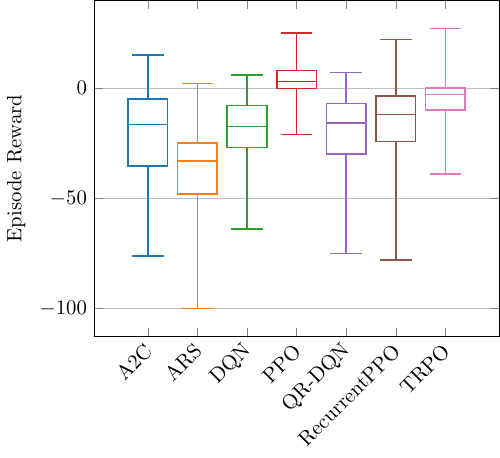}
\caption{Environment: Boxing\\Reward Distribution}
\label{figure:boxingdist}
\end{figure}

\begin{figure}[!ht]
\centering
\includegraphics[width=\linewidth]{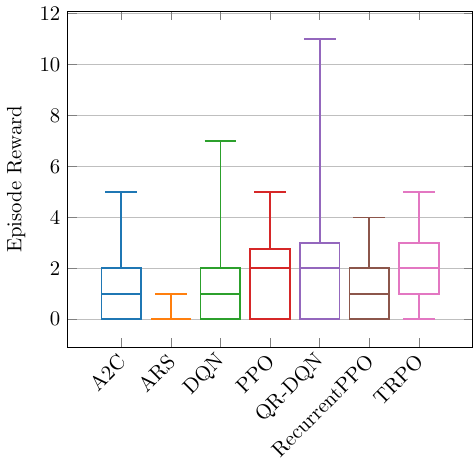}
\caption{Environment: Breakout\\Reward Distribution}
\label{figure:breakoutdist}
\end{figure}

\begin{figure}[!ht]
\centering
\includegraphics[width=\linewidth]{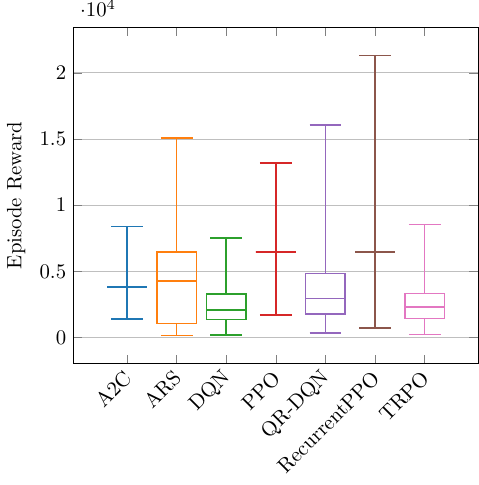}
\caption{Environment: Centipede\\Reward Distribution}
\label{figure:centipededist}
\end{figure}

\begin{figure}[!ht]
\centering
\includegraphics[width=\linewidth]{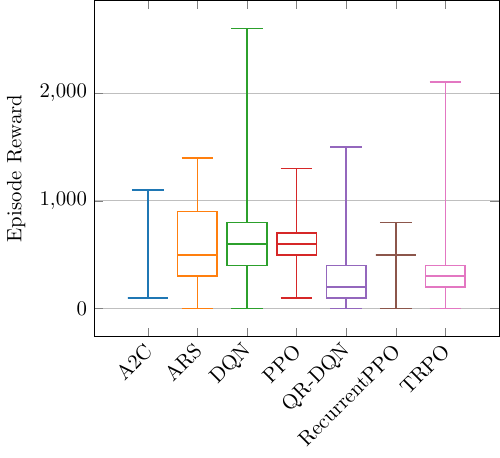}
\caption{Environment: Chopper Command\\Reward Distribution}
\label{figure:choppercommanddist}
\end{figure}

\begin{figure}[!ht]
\centering
\includegraphics[width=\linewidth]{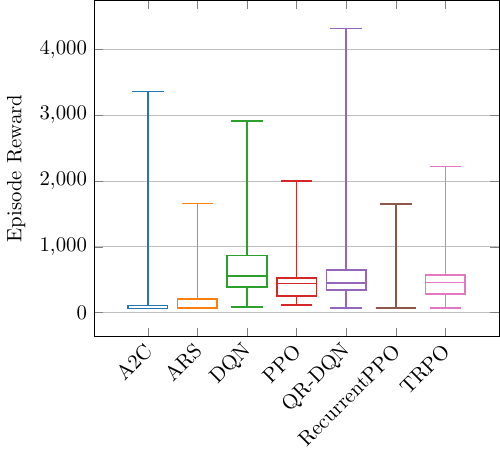}
\caption{Environment: Ms. Pacman\\Reward Distribution}
\label{figure:mspacmandist}
\end{figure}

\begin{figure}[!ht]
\centering
\includegraphics[width=\linewidth]{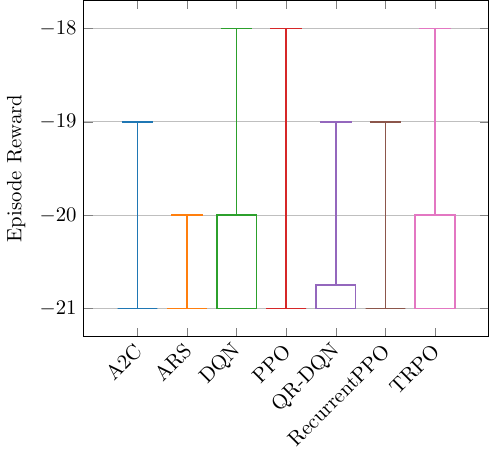}
\caption{Environment: Pong\\Reward Distribution}
\label{figure:pongdist}
\end{figure}

\begin{figure}[!ht]
\centering
\includegraphics[width=\linewidth]{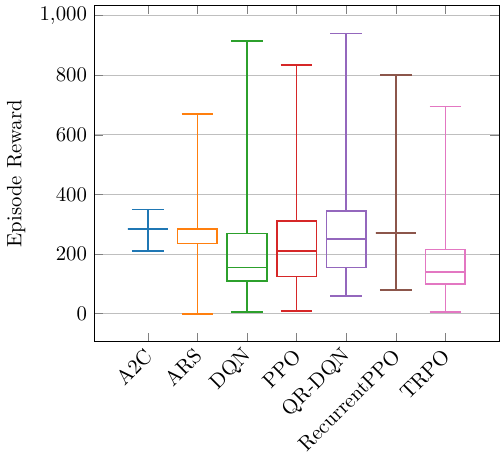}
\caption{Environment: Space Invaders\\Reward Distribution}
\label{figure:spaceinvadersdist}
\end{figure}

\begin{figure}[!ht]
\centering
\includegraphics[width=\linewidth]{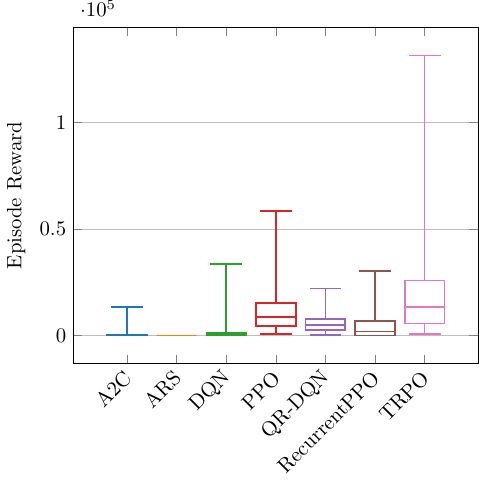}
\caption{Environment: Video Pinball\\Reward Distribution}
\label{figure:videopinballdist}
\end{figure}

To further contextualize energy efficiency, we analyzed the relationship between training time and mean episodic reward across the ten Atari environments. We employed three complementary visualization strategies to characterize training performance. The IQR plots use a rolling-window median with shaded inter-quartile ranges (25-75th percentile) to capture training stability over time while reducing the influence of outliers (Figs. \ref{figure:asteroidsiqr} - \ref{figure:videopinballiqr}). For cases where rolling-window smoothing failed to produce a readable signal, we used binned plots, which apply more aggressive temporal smoothing by aggregating rewards into $20$ time-subdivisions and plotting the median along with standard deviation bands; this provides a clearer view of long-term trends and variability at the cost of finer temporal resolution (Figs. \ref{figure:asteroidsbinned} - \ref{figure:videopinballbinned}). Finally, the box plots abstract away temporal information entirely, instead summarizing the overall distribution of rewards for each algorithm using medians, inter-quartile ranges, and outlier markers, offering a concise but high-level comparison of performance across algorithms (Figs. \ref{figure:asteroidsdist} - \ref{figure:videopinballdist}).

While NPpkWh provides a normalized view of energy efficiency, understanding how quickly algorithms achieve competitive performance is equally critical for practical applications where training time directly correlates with electricity costs and operational feasibility.

As expected, ARS achieved good rewards relatively fast in dense-reward games, reaching stable performance significantly earlier than gradient-based methods. For example, it reached 90\% of its maximum reward at step 112,804 on average, whereas the average across all algorithms for this landmark is step count 113,857. Its derivative-free optimization allowed for rapid exploration of the policy space with fewer computationally expensive gradient updates, making it attractive for time- and cost-constrained deployments. However, in games with sparse or delayed rewards, ARS plateaued early, indicating limited ability to exploit small reward signals.

%\subsection{Figures}\label{app:figures}

RecurrentPPO and TRPO, due to their higher per-update computation, reached 90\% of the maximum reward at steps 172,862 and 228,748, respectively, whereas the average step count was 113,857. Although this is slower than DQN and QR-DQN, we should note that 90\% of maximum rewards for RecurrentPPO and TRPO are moderately higher than that of DQN and QRDQN on average (Table \ref{tab:atari_aggregate}. Thus, both algorithms reached higher average rewards earlier than DQN given the same steps, owing to RecurrentPPO's improved temporal modeling and TRPO's stable trust-region updates. %This faster convergence directly translates into reduced total training time and electricity costs, reinforcing their favorable NPpkWh scores.

In contrast, DQN and QR-DQN exhibited prolonged training phases in the majority of the games, requiring substantially more updates to achieve moderate performance levels. The reliance on large replay buffers and unstable Q-value estimation contributed to inefficient exploration and longer training durations, aligning with their poor energy efficiency.

The box plot distributions (Figs. \ref{figure:asteroidsdist} - \ref{figure:videopinballdist}) further highlight the stability of ARS, which exhibited minimal to no outliers in many games, suggesting consistently stable training dynamics. In contrast, QR-DQN showed a wider spread with frequent outliers, indicating more volatile learning behavior while performing the worst in terms of average NPpkWh. This stability in ARS likely contributes to its superior energy and cost efficiency, as fewer fluctuations in training reduce the need for prolonged exploration and repeated updates.

Overall, these findings highlight that algorithms with better sample efficiency not only consume less energy per kilowatt hour but also complete training faster, providing dual benefits in terms of reduced electricity expenditure and a lower carbon footprint.

%\onecolumn\subsection{Tables}\label{app:tables}
% Remove old tables for improved ones.
%\input{tables/asteroids}
%\input{tables/beamrider}
%\input{tables/boxing}
%\input{tables/breakout}
%\input{tables/centipede}
%\input{tables/chopper_command}
%\input{tables/ms_pacman}
%\input{tables/pong}
%\input{tables/space_invaders}
%\input{tables/video_pinball}

%%%% Win table %%%%%%
%\begin{table}[ht!]
%    \centering
%    \begin{tabular}{ccccc}
%    \toprule
%        Algorithm & Reward & Time & Energy & Emissions\\
%        \midrule
%        A2C & 1 & 0 & 0 & 0\\
%        ARS & 0 & 10 & 9 & 9\\
%        DQN & 1 & 0 & 0 & 0\\
%        PPO & 2 & 0 & 0 & 0\\
%        QR-DQN & 0 & 0 & 1 & 1\\
%        RecurrentPPO & 1 & 0 & 0 & 0\\
%        TRPO & 5 & 0 & 0 & 0\\
%        \bottomrule
%    \end{tabular}
%    \caption{Win summary of the DRL algorithms in each category for ten Atari games.}
%    \label{tab:win_summary}
%\end{table}

\begin{table}[!ht]
\centering
\includegraphics[width=\linewidth]{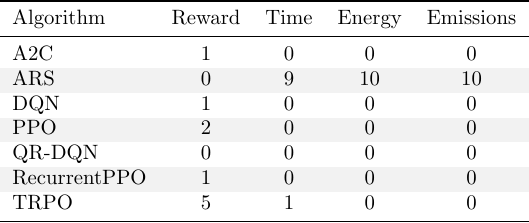}
\caption{Win summary of the DRL algorithms in each category for ten Atari games.}
\label{tab:win_summary}
\end{table}

%%%%%% aggregate results (mean of 10 games) %%%%%
%\begin{table*}[htbp]
%\centering
%\begin{tabular}{lcccccc}
%\toprule
%Algorithm & Reward & Time (s) & Energy (kWh) & Emissions (kgCO2eq) & Cost (Local) & Cost (National) %\\
%\midrule
%A2C & {\cellcolor{red!25}533.68} & 2,060.18 & 14.16 & 10.39 & \$1.56 & \$19.84 \\
%ARS & 599.82 & {\cellcolor{green!25}750.08} & {\cellcolor{green!25}2.83} & {\cellcolor{green!25}%2.08} & {\cellcolor{green!25}\$0.31} & {\cellcolor{green!25}\$3.97} \\
%DQN & 650.12 & 1,379.27 & 11.61 & 8.54 & \$1.28 & \$16.26 \\
%PPO & 2,108.24 & 2,237.98 & 12.45 & 9.16 & \$1.37 & \$17.44 \\
%QR-DQN & 1,078.70 & 1,492.29 & 15.44 & 11.34 & \$1.70 & \$21.63 \\
%RecurrentPPO & 1,273.97 & {\cellcolor{red!25}3,522.10} & {\cellcolor{red!25}23.25} & %{\cellcolor{red!25}17.11} & {\cellcolor{red!25}\$2.56} & {\cellcolor{red!25}\$32.57} \\
%TRPO & {\cellcolor{green!25}2,417.05} & 1,480.26 & 10.96 & 8.05 & \$1.21 & \$15.35 \\
%\bottomrule
%\end{tabular}
%\caption{Mean Aggregate Performance Results Across Atari Games}
%\label{tab:atari_aggregate}
%\end{table*}

\begin{table}[!ht]
\centering
\includegraphics[width=\linewidth]{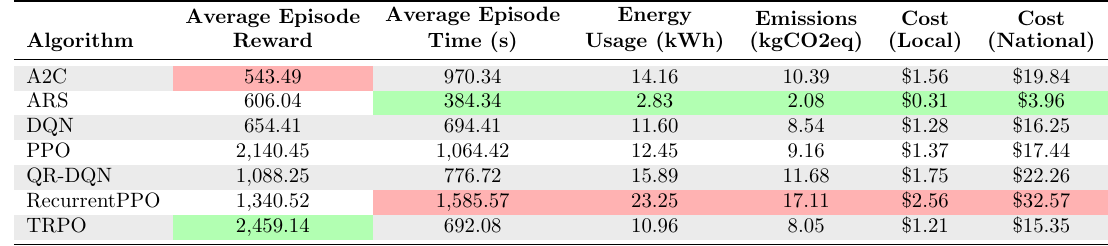}
\caption{Mean Aggregate Performance Results Across Atari Games}
\label{tab:atari_aggregate}
\end{table}

\begin{table}[!ht]
\centering
\includegraphics[width=\linewidth]{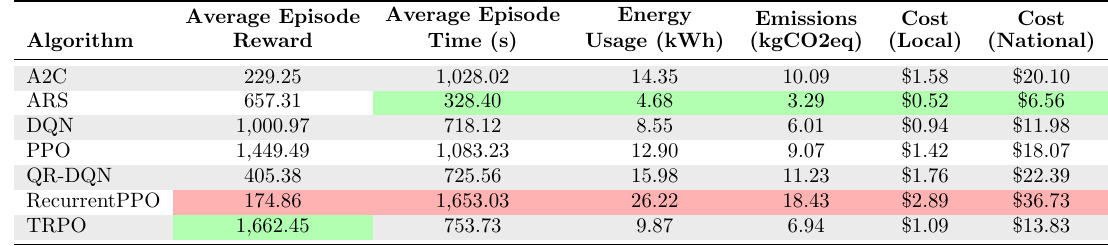}
\caption{Environment: Asteroids}
\label{table:asteroids}
\end{table}

\begin{table}[!ht]
\centering
\includegraphics[width=\linewidth]{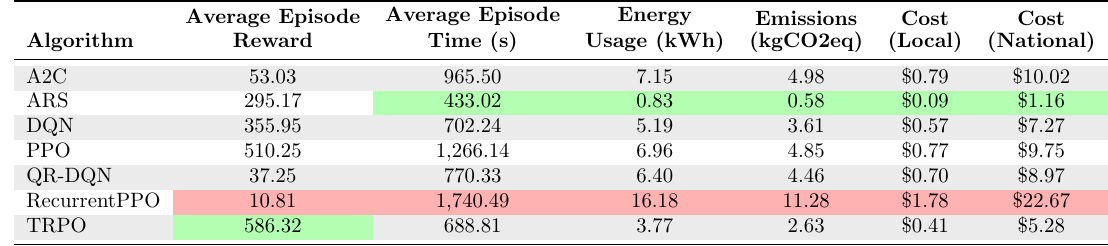}
\caption{Environment: Beam Rider}
\label{table:beamrider}
\end{table}

\begin{table}[!ht]
\centering
\includegraphics[width=\linewidth]{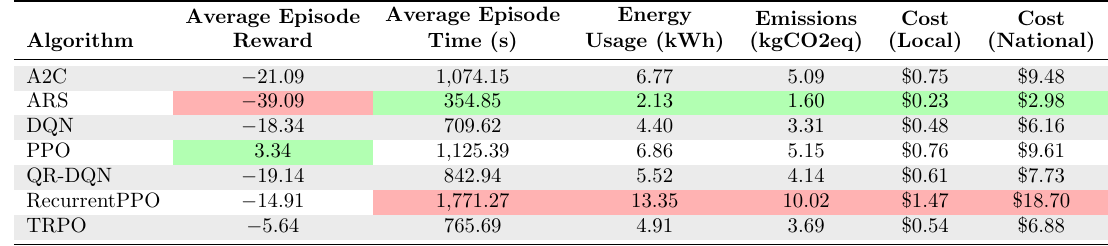}
\caption{Environment: Boxing}
\label{table:boxing}
\end{table}

\begin{table}[!ht]
\centering
\includegraphics[width=\linewidth]{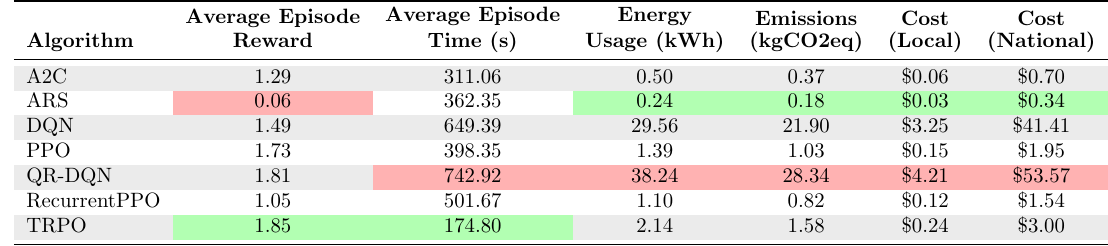}
\caption{Environment: Breakout}
\label{table:breakout}
\end{table}

\begin{table}[!httb]
\centering
\includegraphics[width=\linewidth]{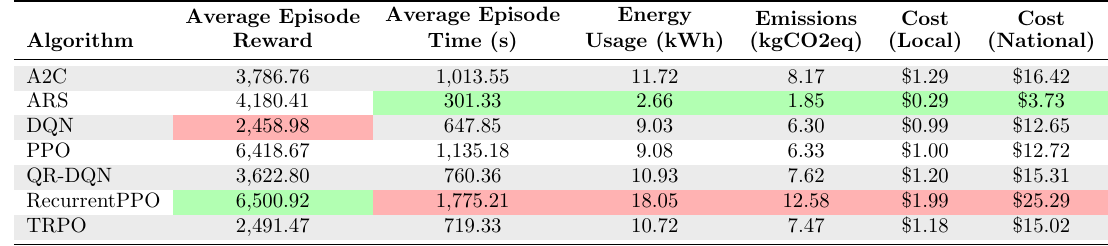}
\caption{Environment: Centipede}
\label{table:centipede}
\end{table}

\begin{table}[!ht]
\centering
\includegraphics[width=\linewidth]{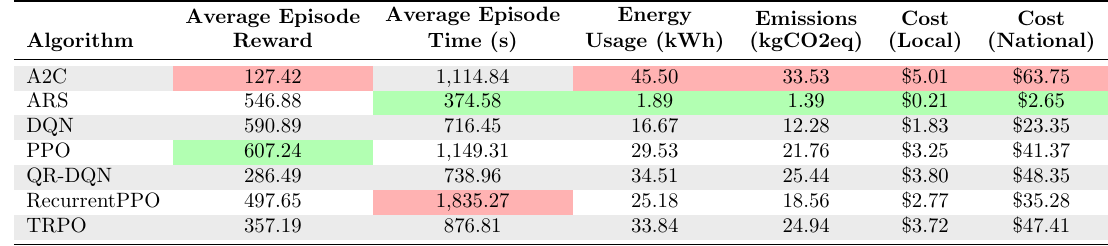}
\caption{Environment: Chopper Command}
\label{table:choppercommand}
\end{table}

\begin{table}[!ht]
\centering
\includegraphics[width=\linewidth]{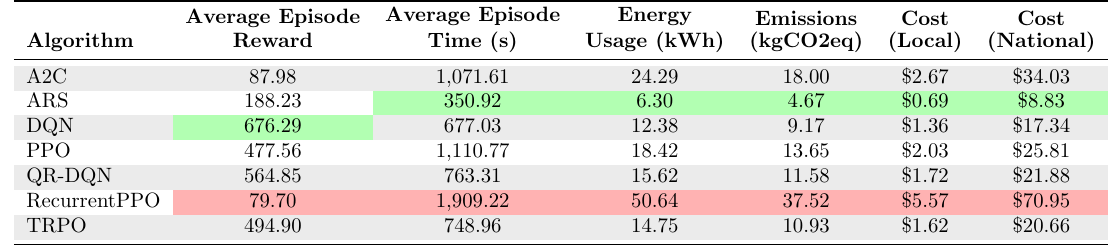}
\caption{Environment: Ms. Pacman}
\label{table:mspacman}
\end{table}

\begin{table}[!ht]
\centering
\includegraphics[width=\linewidth]{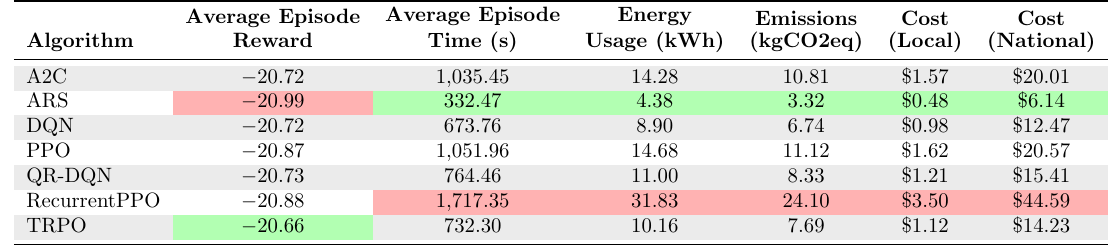}
\caption{Environment: Pong}
\label{table:pong}
\end{table}

\begin{table}[!ht]
\centering
\includegraphics[width=\linewidth]{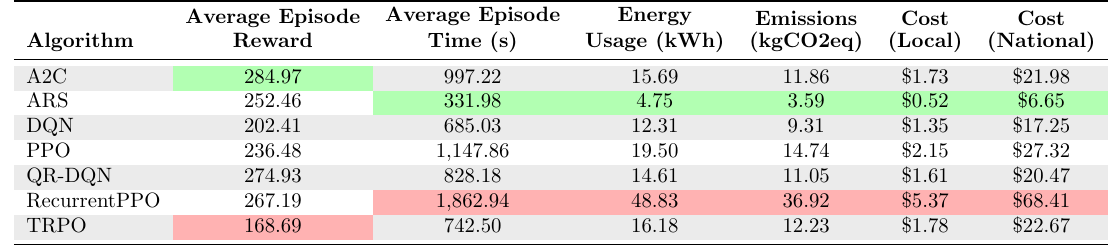}
\caption{Environment: Space Invaders}
\label{table:spaceinvaders}
\end{table}

\begin{table}[!ht]
\centering
\includegraphics[width=\linewidth]{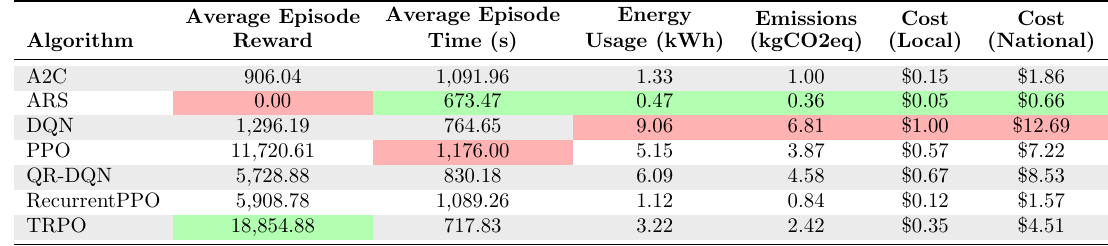}
\caption{Environment: Video Pinball}
\label{table:videopinball}
\end{table}

\subsection*{Electricity Cost Implications}
The practical implications of energy usage can be assessed through electricity cost calculations. Using the U.S. national average electricity price and our local utility rate, ARS consistently emerged as the most economical algorithm (Table \ref{tab:atari_aggregate}). For instance, training ARS for $10^6$ steps on an average Atari game incurred an estimated cost of only \$0.31 (local rate) or \$3.96 (national average), whereas RecurrentPPO cost approximately \$2.56 (local) or \$32.57 (national average) for the same number of steps.  

%TRPO and RecurrentPPO also exhibited favorable cost-efficiency trade-offs. TRPO’s optimized learning dynamics resulted in costs roughly 40-50\% lower than PPO for similar performance levels, while RecurrentPPO reduced training costs by 25-30\%. On the other hand, DQN and QR-DQN, despite their historical dominance in Atari benchmarks, were among the least cost-efficient due to their extensive replay buffer operations.

Across the ten Atari games, we observe consistent trends in energy consumption, emissions, and monetary cost across algorithms. ARS stands out as the most energy-efficient method, requiring $5.20$x less energy than the average energy consumed by the rest, while still achieving competitive rewards in games such as Centipede, Chopper Command, Space Invader, and Beam Rider. In contrast, RecurrentPPO consistently incurs the highest energy usage and carbon emissions, exceeding $1500$ sec. of training time and resulting in costs $8$x higher than ARS on average. Among policy gradient methods, PPO and TRPO generally achieve moderate energy efficiency, whereas QR-DQN occupies the middle range, balancing reasonable performance with moderate energy costs. Notably, the relationship between reward and energy usage is not always proportional. For instance, Breakout and Space Invaders depict cases where QR-DQN and RecurrentPPO achieve high rewards but with disproportionately high energy expenditure, respectively. When we look at the win summary in Table \ref{tab:win_summary}, TRPO is a clear winner in terms of earned rewards, whereas ARS won in all but one game in terms of the lowest energy consumption as well as GHG emission. These findings emphasize that algorithm choice can drastically affect not just learning performance but also energy and cost efficiency, with derivative-free methods like ARS emerging as strong candidates for energy-conscious training.

These cost differences have broader sustainability implications. Large-scale RL training often involves hundreds of millions of steps across many environments, making even small per-step energy savings economically and environmentally significant. For example, scaling ARS's energy savings to 100 million training steps could reduce electricity expenditures by approximately \$2,861 (national average) compared to training a RecurrentPPO agent, alongside preventing the emission of several kilograms of tCO$_2$e depending on regional energy sources.

\subsection*{Limitations and Broader Implications}
No algorithm achieved positive average rewards in games with sparse or delayed rewards (e.g., Boxing and Pong), where all agents scored negative average returns (Tables~\ref{table:boxing} and~\ref{table:pong}) except PPO in Boxing. This suggests that energy efficiency is strongly coupled to environmental reward structure and that algorithms may require environment-specific modifications for optimal performance-per-kWh.

Our study is limited by the use of ten Atari discrete-action games, a single training seed per run, and a fixed hardware configuration. Future work should explore multiple random seeds~\cite{NEURIPS2018_7634EA65}, cross-platform hardware comparisons, and more complex environments, and continuous-action games to further validate these findings. Additionally, hardware-aware algorithm design and carbon-intensity-aware scheduling could further reduce both electricity costs and associated greenhouse gas emissions.

\subsection*{Key Takeaways for Sustainable RL}
The results highlight three critical considerations for energy-aware reinforcement learning:
\begin{itemize}
    \item Derivative-free methods (e.g., ARS) are highly energy and cost-efficient in dense-reward, structured domains.
    \item Sample efficiency outweighs per-step computational cost, as seen with TRPO and PPO, which achieved better NPpkWh despite higher computational complexity per update.
    \item Electricity cost savings scale significantly with large training runs, reinforcing the importance of selecting energy-efficient algorithms for both economic and environmental reasons.
\end{itemize}

\section{Conclusion}
This study systematically quantified the energy consumption and carbon emissions associated with training seven deep reinforcement learning algorithms on a standardized suite of Atari games. Our findings demonstrate that algorithmic choice alone can drive substantial differences in energy efficiency, even under identical hardware and software configurations. Augmented Random Search (ARS) achieved the highest Normalized Performance per kilowatt hour (NPpkWh), outperforming the least efficient algorithm (QRDQN) by more than $4.30$x. These results highlight the potential of derivative-free methods in domains with dense reward signals and structured state representations, where stable performance can be achieved with minimal computational overhead.

PPO and TRPO also showed significant efficiency advantages, achieving $1$-$2.5$x higher NPpkWh compared to other algorithms. In Atari, where stacked frames already provide sufficient state information, the added recurrence in RecurrentPPO introduces computational overhead, and therefore, it did not outperform PPO. TRPO's trust-region updates improved training stability and sample efficiency. Although it incurs higher per-update computational costs, its moderate time to finish the training ultimately translates into lower total energy usage and reduced electricity expenditures. For instance, the energy efficiency of ARS corresponded to an estimated $8.22$x reduction in GHG emissions compared to RecurrentPPO, but only $3.87$x compared to TRPO.

Beyond aggregated NPpkWh metrics, our analysis of reward stability revealed a strong connection between learning dynamics and energy efficiency. Algorithms with narrower inter-quartile ranges and uni-modal reward distributions (e.g., ARS) tended to require fewer redundant training steps, whereas highly variable or multi-modal reward patterns (e.g., QR-DQN) were associated with higher energy consumption. Training-time versus mean-reward trends further reinforced this link, as more stable algorithms reached peak performance in significantly fewer updates.

However, some limitations must be acknowledged. Our experiments focused exclusively on ten Atari environments, which are relatively low-dimensional; performance in tasks with sparser rewards, continuous action spaces, or high-dimensional sensory inputs may yield different efficiency rankings. %Furthermore, only a single random seed and a fixed hardware configuration were used, limiting the statistical generalizability of the results.
Despite these constraints, the implications of this work are clear: careful algorithm selection can meaningfully reduce the carbon footprint and electricity costs of RL research without sacrificing performance. As large-scale AI models continue to drive increasing energy demand, incorporating energy-efficiency metrics such as NPpkWh alongside traditional performance measures offers a practical step toward more sustainable machine learning. Future work should extend this analysis to diverse tasks, hardware platforms, and randomized training runs, while also investigating algorithmic modifications explicitly designed to optimize both performance and energy efficiency.

% \section*{Acknowledgment}
% This work was partly supported by the NSF Cyber-Physical Systems Grants \#1932300 and \#1931767.

%\newpage

\ifCLASSOPTIONcaptionsoff
  \newpage
\fi

%Bibliography
\bibliographystyle{IEEEtran} % We choose the "plain" reference style
\bibliography{refs} % Entries are in the refs.bib file
%\vspace{12pt}


% \end{IEEEbiography}


% % insert where needed to balance the two columns on the last page with
% %\newpage


%\appendix
%\subsection{Tables}\label{app:tables}

%\EOD
\end{document}